\documentclass[letterpaper, 10 pt, conference]{ieeeconf}  

\IEEEoverridecommandlockouts                              

\usepackage{amsmath,graphicx}
\usepackage{booktabs}

\title{\LARGE \bf
Fast Multi-Organ Fine Segmentation in CT Images with Hierarchical Sparse Sampling and Residual Transformer
}

\author{Xueqi Guo, Halid Ziya Yerebakan, Yoshihisa Shinagawa, Kritika Iyer, Gerardo Hermosillo Valadez
\thanks{Xueqi Guo, Halid Ziya Yerebakan, Yoshihisa Shinagawa, Kritika Iyer, and Gerardo Hermosillo Valadez are with Siemens Medical Solutions USA Inc, Malvern, PA, 19355, USA.}
\thanks{Corresponding email:xueqi.guo@siemens-healthineers.com}%
}

\begin{document}

\maketitle
\thispagestyle{empty}
\pagestyle{empty}

\begin{abstract}

Multi-organ segmentation of 3D medical images is fundamental with meaningful applications in various clinical automation pipelines. Although deep learning has achieved superior performance, the time and memory consumption of segmenting the entire 3D volume voxel by voxel using neural networks can be huge. Classifiers have been developed as an alternative in cases with certain points of interest, but the trade-off between speed and accuracy remains an issue. Thus, we propose a novel fast multi-organ segmentation framework with the usage of hierarchical sparse sampling and a Residual Transformer. Compared with whole-volume analysis, the hierarchical sparse sampling strategy could successfully reduce computation time while preserving a meaningful hierarchical context utilizing multiple resolution levels. The architecture of the Residual Transformer segmentation network could extract and combine information from different levels of information in the sparse descriptor while maintaining a low computational cost. In an internal data set containing 10,253 CT images and the public dataset TotalSegmentator, the proposed method successfully improved qualitative and quantitative segmentation performance compared to the current fast organ classifier, with fast speed at the level of $\sim$2.24 seconds on CPU hardware. The potential of achieving real-time fine organ segmentation is suggested.
\newline

\indent \textit{Clinical relevance}— We introduce an innovative fast multi-organ segmentation framework that utilizes hierarchical sparse sampling combined with a Residual Transformer. This approach significantly reduces computation time compared to whole-volume analysis while retaining meaningful hierarchical context through multiple resolution levels. This method enhances both qualitative and quantitative segmentation performance over existing fast organ classifiers, achieving segmentation in approximately 2.24 seconds on standard CPU hardware. This indicates the promising potential for real-time fine organ segmentation in various clinical applications, including scan registration, lesion detection, and landmarking.
\end{abstract}

\section{INTRODUCTION}
Multi-organ segmentation in computed tomography (CT) images has been a foundation of a variety of computer-assisted diagnostic systems and the automation of various clinical workflows. Segmenting organs of interest, at risk, or involved in diagnosis and treatment is crucial in the planning of radiation therapies, surgeries, and image guidance systems \cite{gibson2018automatic}, with the desired run time at the level of seconds. Thus, it is of great interest to have fast and accurate algorithms to segment organs in medical images.

Recent developments in deep learning have gained success in achieving multi-organ segmentation. The structure of U-Net \cite{ronneberger2015u} and the 3-D variations \cite{milletari2016v} have been widely deployed in image segmentation tasks in the medical \cite{gibson2018automatic} and natural image domains, with the skip connections having the capability of integrating both local and global contexts of features. However, convolutional networks have limitations in limited reception fields. Transformers and its variations \cite{hatamizadeh2022unetr,shaker2024unetr} have been introduced to analyze the entire field of view with a multi-head attention mechanism, but the complexity of the model and the cost of computation could introduce significant challenges in efficiency. The combination of convolutional networks and Transformers has been investigated in object detection of natural images \cite{carion2020end} and medical anomalies \cite{ramezani2024lung}, but the detection box might not give sufficiently accurate voxel-wise segmentation masks, especially for the edges. Mamba \cite{gu2023mamba} was proposed as a selective state space model to address the efficiency problem in Transformer, but the long context dependency might not align perfectly in the medical image segmentation domain that requires precise and localized details.

In multi-organ segmentation tasks, voxel-level computation is generally slow. The inference of the segmentation method based on nn-UNet \cite{isensee2021nnu} in the TotalSegmentator CT dataset \cite{wasserthal2023totalsegmentator} takes up to 3 minutes 32 seconds on a GPU. The efficiency of Transformer-based methods has been reported to reach a run time speed of $\sim$60 seconds on a GPU \cite{shaker2024unetr}. Real-time-level computational efficiency has not yet been achieved, especially on CPU-only hardware. To address the need for faster computation and runtime, object detection-based methods including organ bounding boxes \cite{xu2019efficient} and landmark matching \cite{yerebakan2023hierarchical} have been investigated as real-time alternatives with a fast speed of $\sim$0.25 seconds, but the fast speed and coarse estimates compensate for computation accuracy, which might not be applicable in tasks requiring refined boundaries. A classifier-based segmentation model was proposed to achieve fast real-time-level segmentation \cite{yerebakan2024real}. This model achieves approximately 5 seconds for coarse segmentation and an additional 9 seconds for edge refinement. However, the classifier operates at a coarse block level rather than voxel-level resolution, only returning the organ class of one coarse block instead of the voxel-level predictions. This requires further edge refinement to achieve the desired precision in segmentation tasks.

\begin{figure*}[!t]
\centering
\includegraphics[width=\textwidth]{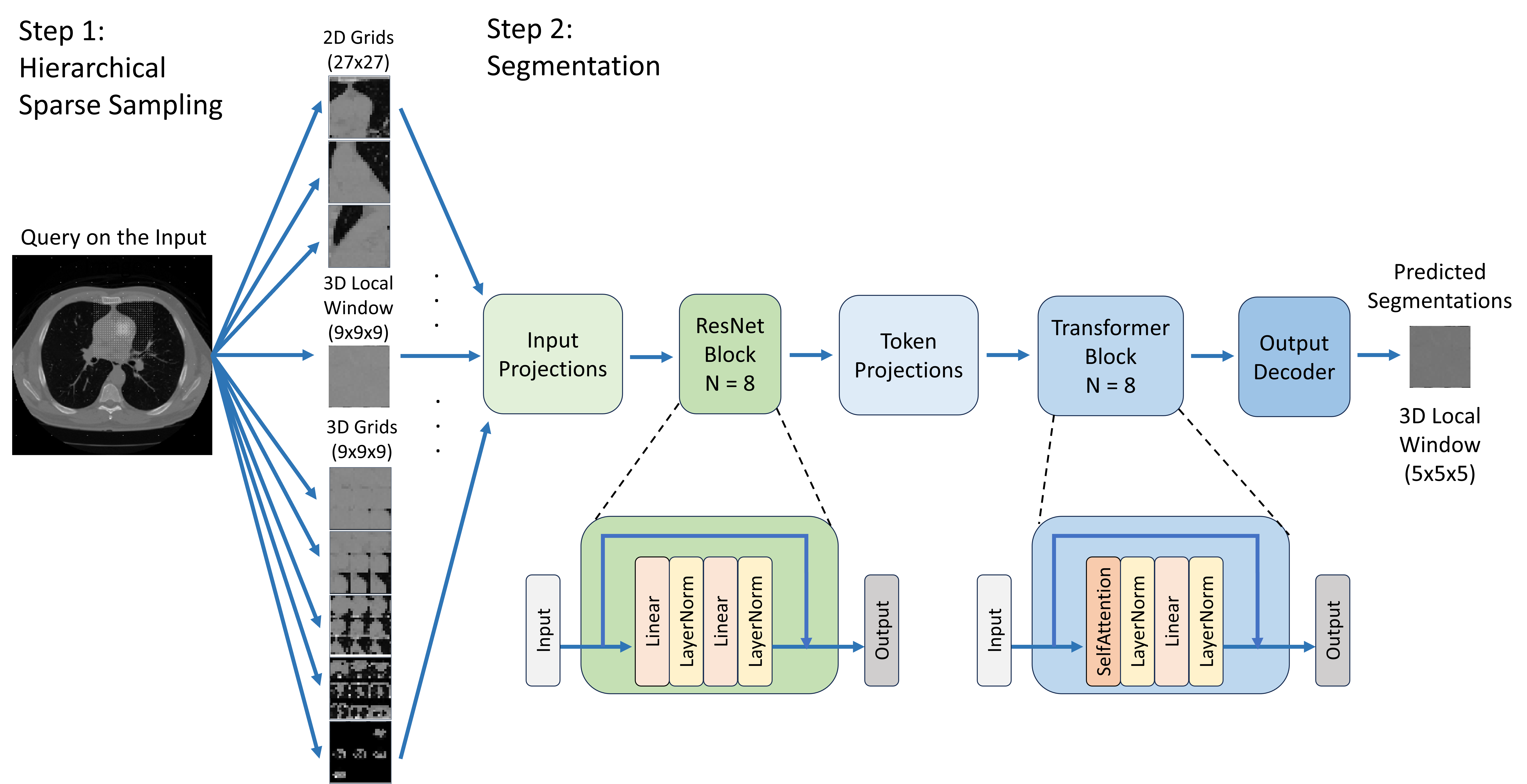}
\caption{The architecture of the proposed segmentation method.} 
\label{network}
\end{figure*}

In this work, we propose a fast fine segmentation framework with the usage of a hierarchical sparse sampling strategy and a Residual Transformer network returning high resolution segmentation masks. The sampling strategy allows the network to parse hierarchical information with an enlarged reception field under reduced data. The structure of the Residual Transformer allows the model to more efficiently extract and fuse information from multiple resolution levels from the sparse sampling while minimizing model complexity. By querying each point on the grid, the full volume segmentation can be reconstructed in seconds on a CPU. The proposed method effectively enhanced both qualitative and quantitative segmentation performance while maintaining a fast processing speed.

\section{METHODS}

Figure \ref{network} shows the workflow of the proposed segmentation method. Hierarchical sparse sampling is first implemented in the query voxel to generate sparse descriptors extracting 2-D and 3-D grids across multiple resolutions, and then a Residual Transformer was applied to decode and predict the voxel-level segmentations of the local grid block. 

\subsection{Hierarchical Sparse Sampling Strategy}
\begin{figure*}[!t]
\centering
\includegraphics[width=0.8\textwidth]{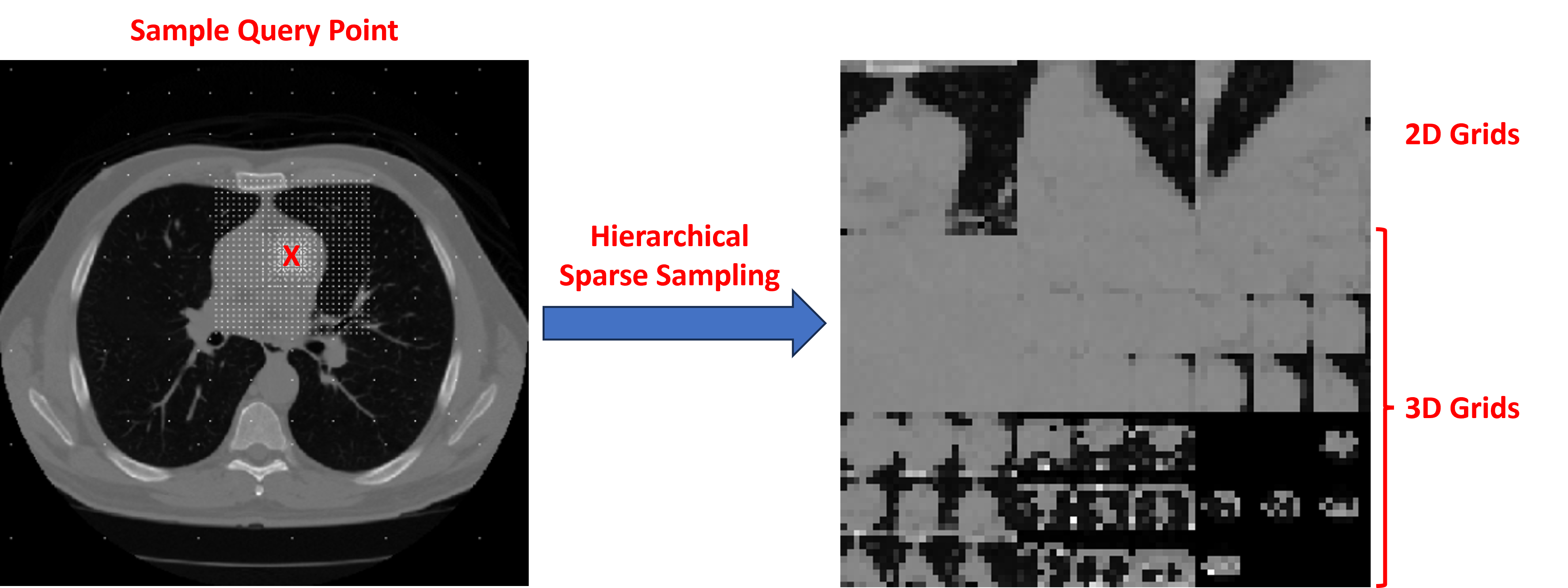}
\caption{The details of the hierarchical sparse sampling strategy demonstrating the descriptor generated from the sample query point on the 3D volume. In the figure, the red ``X'' is the query location from where generates the descriptor through sparse sampling. The white dots demonstrates the sampling location in multiple spatial resolutions.} 
\label{sampling}
\end{figure*}

Figure \ref{sampling} demonstrates the details of the hierarchical sparse sampling strategy. The hierarchical sparse sampling strategy was proposed to mitigate the huge computational need for voxel-wise segmentation of the entire volume while also capturing the anatomical context from a large field of view \cite{yerebakan2023hierarchical,yerebakan2024real}. Given the consistency of human anatomy, similar locations produce analogous descriptors. To enhance sampling, we employ multiple regular grids at various resolutions, allowing us to hierarchically cover larger areas. 

During execution, the descriptor computation is optimized for memory lookups, where memory locations are calculated by adding offsets to the current voxel. Fixed offsets were given through the hierarchical sparse sampling procedure, generating a descriptor of the query location that includes both 2-D and 3-D grids at multiple resolutions with hierarchical information with a dimension of 9 × 9 × 9 × 9. The first three 2-D grids in the sampled descriptor are three 27 × 27 orthogonal planes at a resolution of 4 mm. The following six 3-D grids are six 9 × 9 × 9 grids at multiple resolutions of 2, 3, 5, 12, 28, and 64 mm, respectively, from fine to coarse. This spacing resolution setting helps avoid overlapping samples across different resolution grids. These 2-D and 3-D descriptors can be reconstructed and visualized as an 81 x 81 2-D image, as shown in Figure \ref{sampling}, by placing each 27 x 27 block in nine positions. The total dimensionality of the sampled descriptor is 6561. Through this sparse sampling strategy, hierarchical information is obtained from not only the local region but also the global context, effectively extracting information from a large receptive field with reduced data.

\subsection{Residual Transformer Segmentation Network}
We propose to use the structure of a Residual Transformer to generate segmentation masks from sparsely sampled descriptors. The nine grids of the descriptor were first flattened and projected independently to extract information from multiple resolutions while preserving essential features. Each independent projection layer for each grid in the descriptor has a hidden size of 32, allowing for a compact representation. Afterward, the nine projections were concatenated and processed through a linear layer with a hidden size of 144, enhancing the model's ability to integrate features from different grids.

Next, a series of residual blocks fuse and extract meaningful information from the input projections, utilizing a combined two-layer linear feedforward network followed by layer normalization to generate feature representations as the input tokens for the subsequent Transformer layers. Each linear layer maintains a hidden size of 144, supporting the retention of complex relationships within the data. After the token embedding layer, a sequence of Transformer encoder layers with the architecture of multi-head self-attention and a feed-forward network captures the contextual information among the transformed tokens. Each Transformer encoder layer employs the same feed-forward dimension and two heads in the multi-head self-attention mechanism, enabling the model to simultaneously focus on different parts of the input that are originally from multiple resolutions. Residual connections are also incorporated in each Transformer encoder block to concatenate feature maps at every level, facilitating gradient flow and enhancing model performance.

Before reaching the output decoder, the extracted feature map undergoes a final linear layer that consolidates information and reshapes it to a size of 9 × 9 × 9 × 8. This is then concatenated with the 9 × 9 × 9 grid of the 3-D local window, which provides local fine details that are essential for accurate segmentation. The feature map is decoded through two convolutional layers with a kernel size of 3 × 3 × 3, aligning the output dimension with the segmentation mask that covers the local grid block centered around the query voxel. The first convolutional layer consists of 9 kernels, while the second output convolutional layer has the number of kernels that match the number of organ classes in the dataset, with a stride of 2 to downsample the output effectively. Instead of employing a classifier that predicts only the label at the query point, our segmentation network predicts the segmentation of a central local 3-D window with dimensions of 5 × 5 × 5 for each query. This approach significantly improves prediction efficiency compared to single-voxel classifiers \cite{yerebakan2024real}, as it captures spatial context and anatomical relationships, leading to more accurate and comprehensive segmentation masks. This method can be potentially adapted for various clinical applications, including automated organ delineation and diagnosis in medical imaging, enhancing both speed and precision.

\subsection{Model Training and Whole Volume Segmentation}
We trained and evaluated the fast segmentation model on an internal dataset that contains 10,253 CT images with 119 organ classes and the public dataset TotalSegmentator \cite{wasserthal2023totalsegmentator}. This study was performed following the principles of the Declaration of Helsinki, approved by Siemens Ethics Committee. We randomly split the internal dataset patient-wise with a train/test ratio of 9:1 and followed the official split of the TotalSegmentator. For training, we generated sparse descriptors that were sampled from random locations both globally and from each class of organ to achieve balanced sampling. The segmentation label is derived from the 5 × 5 × 5 local grid of the center voxel of the mask. We randomly sampled 1,000 descriptors per training image, with 10\% sourced from the balanced set. A random test subset containing 100 test subjects was selected to generate evenly sampled sparse descriptors to evaluate the segmentation performance of the whole volume. The model was trained using cross-entropy loss using an Adam optimizer (learning rate=3e-4, weight decay=1e-5) and evaluated on an NVIDIA A100 GPU.

After the model has been successfully trained, we can systematically query the volume at even intervals of 10 mm grids to reconstruct the segmentation of the whole volume. Compared to an organ classifier \cite{yerebakan2024real} that requires querying every location in the image or utilizing edge refinement, this fast segmentation network could return voxel-wise segmentation labels for each 10-mm block, obtaining high-resolution segmentation predictions within a single query in real-time. The CPU run-time speed of the whole volume segmentation was tested on a workstation with Intel Core i7-12850HX Processor.

\section{RESULTS}

\subsection{Ablative Studies}

\newcommand{\tabincell}[2]{\begin{tabular}{@{}#1@{}}#2\end{tabular}}
\begin{table*}[!h]
    \centering
    \hspace{5pt}
    \caption{The mean dice scores on the test set from the ablative studies of the backbone selection of the fast segmentation network, with the best result \textbf{in bold}.}
    \resizebox{0.6\textwidth}{!}{
    \begin{tabular}{cccc}
        \toprule
         & Internal & TotalSegmentator & Whole Volume \\
        \midrule
        U-Net    & 0.501  & 0.265 & 0.425 \\
        ResNet    & 0.777  & 0.621 & 0.710 \\
        Transformer    & 0.719  & 0.490 & 0.316 \\
        Mamba   & 0.750  & 0.584 & 0.680 \\
        ResNet+Mamba   & 0.780  & 0.688 & 0.716 \\
        \tabincell{c}{ResNet+Transformer \\(Proposed)}    & \textbf{0.784}  & \textbf{0.721} & \textbf{0.720} \\
        \bottomrule
    \end{tabular}
    }
    \label{ablative}
\end{table*}

The effectiveness of the hierarchical sparse sampling strategy has been demonstrated in current studies \cite{yerebakan2023hierarchical,yerebakan2024real}. Here, we comprehensively evaluated the backbone selection and alternative structures of the segmentation model as ablative studies. The dice scores on the descriptors from internal and public datasets, as well as the evaluation of the whole volume segmentation, are reported in Table \ref{ablative}. Among the four backbones, ResNet is able to achieve a higher dice score than U-Net, Transformer or Mamba, possibly due to its suitability to analyze hierarchical data. In the whole volume evaluation set, the performances of the Transformer and Mamba decreased more than those of the ResNet, possibly due to the dependencies of long-term and spatial information. Concatenating ResNet with the Transformer further enhances the contextual awareness of the model and provides more meaningful feature representations with the Transformer input tokens for analysis across multiple resolutions within the sparse descriptors. Thus, the proposed method successfully achieved the highest dice scores in all three settings.

\subsection{Visualization of Whole Volume Segmentation}

\begin{figure*}[!t]
\centering
\includegraphics[width=0.95\textwidth]{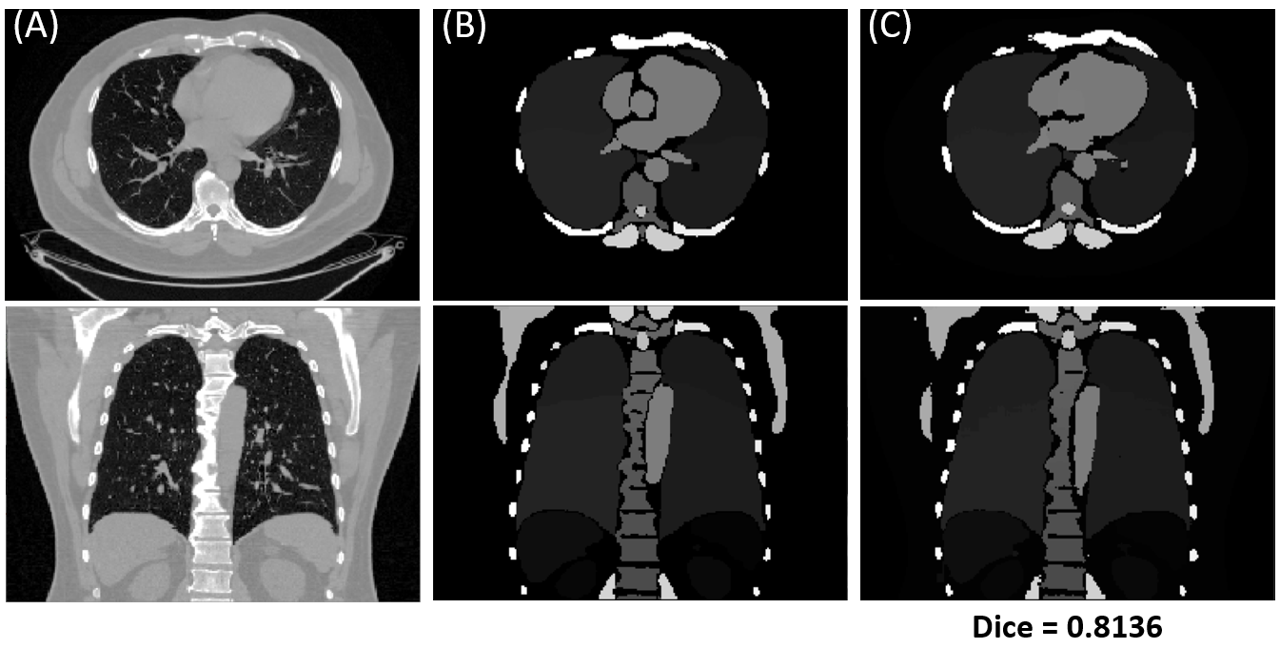}
\caption{Visualization of a sample whole volume segmentation result of the proposed method, with the whole volume multi-class dice score annotated. (A) The CT image; (B) Segmentation ground truth labels; (C) Segmentation result from the proposed method.} 
\label{segresult}
\end{figure*}

Sample whole volume segmentation result of the proposed method is visualized in Figure \ref{segresult}, with the dice score annotated. The proposed method is capable of successfully reconstructing meaningful whole-volume segmentation results despite being trained using only random sparse samples, with fine edge details and clear organ boundaries. Though the 3-D whole volume segmentations were reconstructed from the results of 2-D slices, in the coronal view, the 3-D segmentation masks are smooth with realistic boundaries, without obvious stitching artifacts or slice inconsistencies.

\subsection{Inference Time Comparison}

\begin{table}[h]
    \centering
    \caption{The inference time of the proposed method in different settings.}
    \resizebox{0.4\textwidth}{!}{
    \begin{tabular}{cc}
        \toprule
        Dataset and Hardware & Inference time (s) \\
        \midrule
        Internal (GPU) & 12.00 \\
        TotalSegmentator (GPU) & 2.59 \\
        Whole Volume (CPU) & 2.24 \\
        \bottomrule
    \end{tabular}
    }
    \label{inference}
\end{table}

Table \ref{inference} summarizes the inference time of the proposed fast segmentation method. For GPU evaluations based on balanced random samples across all the images, the proposed method achieved a total inference time of $\sim$12 seconds on the internal evaluation set that included 960 subjects and $\sim$2.59 seconds on the public TotalSegmentator test set. We also compared the conventional voxel-wise segmentation method with the sparse segmentator.  We trained and evaluated the traditional nnUNet-based method using the public TotalSegmentator dataset. Though this nnUNet-based method was able to achieve the best dice score of 0.921, the total evaluation time was $\sim$11 minutes.

The proposed method achieved an average CPU inference time for segmenting a whole CT volume of $\sim$ 2.24 seconds. This is more than four times faster compared to the current fast segmentation framework utilizing a grid point classifier and edge refinement that was reported to have an average run time of 9.51$\pm$2.72 seconds \cite{yerebakan2024real}. The nnUNet-based method was reported to have a runtime of 1-3 minutes per subject on GPU hardware \cite{wasserthal2023totalsegmentator}, and the runtime of segmenting one test subject was $\sim$26 seconds on our in-house NVIDIA A100 GPU. The proposed method is closer to achieving real-time fine segmentation for multi-organ tasks without GPU hardware requirements.

\section{CONCLUSIONS}
In this work, we propose a novel fast multi-organ fine segmentation framework with the usage of a hierarchical sparse sampling strategy and a Residual Transformer network returning high-resolution segmentation masks. The proposed method successfully overcame the limitation of the current classifier-based fast segmentation method that includes querying each location of the image and requires two-step edge refinement, effectively improving both qualitative and quantitative segmentation performance compared to the current fast organ classifier, with a fast whole-volume inference speed at the level of $\sim$2.24 seconds on a CPU. The potential of using this framework to accelerate organ segmentation and various clinical applications is further suggested, including scan registration, lesion detection, and landmarking. Future work includes evaluating generalization between different scanners or institutions, investigating other efficient strategies for hierarchical sampling and network structures to better utilize the global anatomical feature context, as well as the combination with efficient medical foundation models \cite{bae2024samu}, weakly supervised medical image segmentation \cite{guo2023seam}, and unsupervised detection of medical abnormalities \cite{pinaya2022fast}.






\section*{ACKNOWLEDGMENT}
Xueqi Guo, Halid Ziya Yerebakan, Yoshihisa Shinagawa, Kritika Iyer, and Gerardo Hermosillo Valadez are Siemens Healthineers employees. No funding was received. The authors have no other conflict of interest to disclose.


\bibliographystyle{IEEEbib}
\bibliography{refs}

\begin{thebibliography}{10}

\bibitem{gibson2018automatic}
Eli Gibson, Francesco Giganti, Yipeng Hu, Ester Bonmati, Steve Bandula, Kurinchi Gurusamy, Brian Davidson, Stephen~P Pereira, Matthew~J Clarkson, and Dean~C Barratt,
\newblock ``Automatic multi-organ segmentation on abdominal ct with dense v-networks,''
\newblock {\em IEEE transactions on medical imaging}, vol. 37, no. 8, pp. 1822--1834, 2018.

\bibitem{ronneberger2015u}
Olaf Ronneberger, Philipp Fischer, and Thomas Brox,
\newblock ``U-net: Convolutional networks for biomedical image segmentation,''
\newblock in {\em Medical image computing and computer-assisted intervention--MICCAI 2015: 18th international conference, Munich, Germany, October 5-9, 2015, proceedings, part III 18}. Springer, 2015, pp. 234--241.

\bibitem{milletari2016v}
Fausto Milletari, Nassir Navab, and Seyed-Ahmad Ahmadi,
\newblock ``V-net: Fully convolutional neural networks for volumetric medical image segmentation,''
\newblock in {\em 2016 fourth international conference on 3D vision (3DV)}. Ieee, 2016, pp. 565--571.

\bibitem{hatamizadeh2022unetr}
Ali Hatamizadeh, Yucheng Tang, Vishwesh Nath, Dong Yang, Andriy Myronenko, Bennett Landman, Holger~R Roth, and Daguang Xu,
\newblock ``Unetr: Transformers for 3d medical image segmentation,''
\newblock in {\em Proceedings of the IEEE/CVF winter conference on applications of computer vision}, 2022, pp. 574--584.

\bibitem{shaker2024unetr}
Abdelrahman~M Shaker, Muhammad Maaz, Hanoona Rasheed, Salman Khan, Ming-Hsuan Yang, and Fahad~Shahbaz Khan,
\newblock ``Unetr++: delving into efficient and accurate 3d medical image segmentation,''
\newblock {\em IEEE Transactions on Medical Imaging}, 2024.

\bibitem{carion2020end}
Nicolas Carion, Francisco Massa, Gabriel Synnaeve, Nicolas Usunier, Alexander Kirillov, and Sergey Zagoruyko,
\newblock ``End-to-end object detection with transformers,''
\newblock in {\em European conference on computer vision}. Springer, 2020, pp. 213--229.

\bibitem{ramezani2024lung}
Hooman Ramezani, Dionne Aleman, and Daniel L{\'e}tourneau,
\newblock ``Lung-detr: Deformable detection transformer for sparse lung nodule anomaly detection,''
\newblock {\em arXiv preprint arXiv:2409.05200}, 2024.

\bibitem{gu2023mamba}
Albert Gu and Tri Dao,
\newblock ``Mamba: Linear-time sequence modeling with selective state spaces,''
\newblock {\em arXiv preprint arXiv:2312.00752}, 2023.

\bibitem{isensee2021nnu}
Fabian Isensee, Paul~F Jaeger, Simon~AA Kohl, Jens Petersen, and Klaus~H Maier-Hein,
\newblock ``nnu-net: a self-configuring method for deep learning-based biomedical image segmentation,''
\newblock {\em Nature methods}, vol. 18, no. 2, pp. 203--211, 2021.

\bibitem{wasserthal2023totalsegmentator}
Jakob Wasserthal, Hanns-Christian Breit, Manfred~T Meyer, Maurice Pradella, Daniel Hinck, Alexander~W Sauter, Tobias Heye, Daniel~T Boll, Joshy Cyriac, Shan Yang, et~al.,
\newblock ``Totalsegmentator: robust segmentation of 104 anatomic structures in ct images,''
\newblock {\em Radiology: Artificial Intelligence}, vol. 5, no. 5, 2023.

\bibitem{xu2019efficient}
Xuanang Xu, Fugen Zhou, Bo~Liu, Dongshan Fu, and Xiangzhi Bai,
\newblock ``Efficient multiple organ localization in ct image using 3d region proposal network,''
\newblock {\em IEEE transactions on medical imaging}, vol. 38, no. 8, pp. 1885--1898, 2019.

\bibitem{yerebakan2023hierarchical}
Halid~Ziya Yerebakan, Yoshihisa Shinagawa, Mahesh Ranganath, Simon Allen-Raffl, and Gerardo~Hermosillo Valadez,
\newblock ``A hierarchical descriptor framework for on-the-fly anatomical location matching between longitudinal studies,''
\newblock in {\em International Workshop on Lesion Evaluation and Assessment with Follow-Up}. Springer, 2023, pp. 59--68.

\bibitem{yerebakan2024real}
Halid~Ziya Yerebakan, Yoshihisa Shinagawa, and Gerardo~Hermosillo Valadez,
\newblock ``Real time multi organ classification on computed tomography images,''
\newblock {\em International Workshop on Data Engineering in Medical Imaging}, 2024.

\bibitem{bae2024samu}
Joseph Bae, Xueqi Guo, Halid Yerebakan, Yoshihisa Shinagawa, and Sepehr Farhand,
\newblock ``Samu: An efficient and promptable foundation model for medical image segmentation,''
\newblock in {\em International Workshop on Foundation Models for General Medical AI}. Springer, 2024, pp. 134--142.

\bibitem{guo2023seam}
Xueqi Guo, Mohamad Abdi, Yoshihisa Shinagawa, Anna Jerebko, and Sepehr Farhand,
\newblock ``Seam-stress: A weakly supervised framework for interstitial lung disease segmentation in chest ct,''
\newblock in {\em 2023 IEEE 20th International Symposium on Biomedical Imaging (ISBI)}. IEEE, 2023, pp. 1--4.

\bibitem{pinaya2022fast}
Walter~HL Pinaya, Mark~S Graham, Robert Gray, Pedro~F Da~Costa, Petru-Daniel Tudosiu, Paul Wright, Yee~H Mah, Andrew~D MacKinnon, James~T Teo, Rolf Jager, et~al.,
\newblock ``Fast unsupervised brain anomaly detection and segmentation with diffusion models,''
\newblock in {\em International Conference on Medical Image Computing and Computer-Assisted Intervention}. Springer, 2022, pp. 705--714.

\end{thebibliography}

\end{document}